# Non-Contact NIR PPG Sensing through Large Sequence Signal Regression

Timothy Hanley, Dara Golden, Robyn Maxwell, Joseph Lemley, and Ashkan Parsi

*OCTO Sensing Team, Xperi Inc., Galway, Ireland*

**Abstract**

Non-Contact sensing is an emerging technology with applications across many industries from driver monitoring in vehicles to patient monitoring in healthcare. Current state-of-the-art implementations focus on RGB video, but this struggles in varying/noisy light conditions and is almost completely unfeasible in the dark. Near Infra-Red (NIR) video, however, does not suffer from these constraints. This paper aims to demonstrate the effectiveness of an alternative Convolution Attention Network (CAN) architecture, to regress photoplethysmography (PPG) signal from a sequence of NIR frames. A combination of two publicly available datasets, which is split into train and test sets, is used for training the CAN. This combined dataset is augmented to reduce overfitting to the 'normal' 60 – 80 bpm heart rate range by providing the full range of heart rates along with corresponding videos for each subject. This CAN, when implemented over video cropped to the subject's head, achieved a Mean Average Error (MAE) of just 0.99 bpm, proving its effectiveness on NIR video and the architecture's feasibility to regress an accurate signal output.

## 1  Introduction

Non-contact sensing is the act of obtaining an individual's health signals, without any hardware, etc. being physically in contact with them. This is usually achieved using cameras to detect changes or motions often imperceptible to the human eye, that can be regressed to obtain the desired health metric. In the case of PPG, there are slight changes in colour to the skin, caused by blood rushing to and from the heart [Wu et al., 2012]. These colour changes are detectable in both NIR and RGB video, however, they are more pronounced in RGB.

This has enormous potential across multiple sectors from applications in the health industry to in-cabin driver monitoring systems (DMS). There are health situations where it may be uncomfortable for the subject to 'wear' the sensors, or it may be the case that it is simply unfeasible to deploy a contact-based sensor, such as in a DMS. In a health setting, NIR can operate in the dark, to allow for monitoring of the patient's heart rate throughout the night or when they are sleeping, with no discomfort. In a DMS, NIR, especially in the range of 940nm, provides substantial reductions in noise in comparison to RGB, reducing the noise produced by external and uncontrollable factors [Magdalena Nowara et al., 2018]. The use of NIR cameras, along with suitable NIR illuminators, can offset some of the problems encountered in these scenarios.

Xperi's research group proposes a method to accurately calculate heart rate by means of regressing a PPG signal from NIR video. This method consists of a CAN architecture that predicts a large sequence of PPG signals, given a large sequence [Liu et al., 2020] of NIR frames as inputs.

## 2 Methodology

### 2.1 Model

The model utilises a CAN architecture, heavily influenced by DeepPhys [Chen and McDuff, 2018], with the final layer adjusted to predict N signal samples. This adjusted layer also employs the Snake activation function [Ziyin et al., 2020], to improve the model's capability to learn the semi-periodic signal. When regressing a signal where the length of the PPG sample is greater than N, the inference is run for every N sample sequence, with the common outputs averaged to produce the signal waveform.

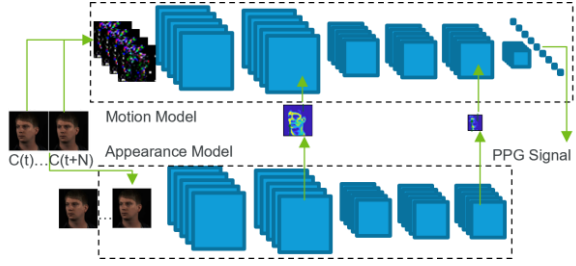

**Figure 1: CAN architecture [Chen and McDuff, 2018]**

|  | MR-NIRP (Indoor) | MR-NIRP (Driving) |
|---|---|---|
| **No. of Subjects** | 8 | 19 |
| **No. of Videos** | 15 | 190 |
| **NIR Wavelengths** | 940 nm | 940 nm, 975 nm |
| **Image Dimension** | 640 x 640 | 640 x 640 |
| **Scenarios** | Indoor | Garage (Indoor), Driving |
| **Motion Levels** | Still, Small | Still, Small, Large |

**Table 1: MR-NIRP datasets outline**

### 2.2 Dataset

The dataset used for training and testing purposes is a combination of both publicly available MR-NIRP datasets produced by the Rice Computational Imaging Lab [Magdalena Nowara et al., 2018] [Nowara et al., 2020].

#### 2.2.1 Dataset Corrections

Due to discrepancies in the dataset, such as an inconsistent PPG sampling frequency and dropped frames, much of the initial work was focused on correcting these. The varying sampling frequency of the PPG ground truth signal was rectified by considering the dropping of samples at the buffer and interpolating those missing samples. Further, any videos/portions of videos where the frames/signals could not be verified were removed for the purpose of this experiment.

|  | MR-NIRP Augmented |
|---|---|
| **No. of Subjects** | 26 |
| **No. of Videos** | 2079 (Augmented & Original) |
| **NIR Wavelengths** | 940 nm, 975 nm |
| **Image Dimension** | 64 x 64 |
| **Scenarios** | Indoor, Garage (Indoor), Driving |
| **Motion Levels** | Still, Small, Large |
| **Heart Rate Ranges** | 40 – 140 bpm |

**Table 2: Augmented and combined MR-NIRP dataset**

#### 2.2.2 PPG Normalisation

To prevent the model from encountering issues learning the DC component of the PPG signal, these signals were normalised between 1 and 0, in such a way that each peak is a 1, and each trough is a 0. The theory behind this is that the model may, given a large sequence (64 samples, ≈ 2 seconds), find it easier to locate the peaks in the sequence rather than detect and quantify the increase/decrease in the signal.

#### 2.2.3 Heart Rate Augmentation

Initial work showed that the model was liable to overfit to the average heart rate range of the dataset, which in this case was discovered to be 60 – 80 bpm. Therefore, to ensure a broader range of heart rates are successfully detected, the dataset was augmented to provide an equal distribution of heart rates in the 40 – 140 bpm range.

This augmentation is achieved by effectively 'stretching' or 'squeezing' the signals and videos with samples interpolated to create an effective 30 fps video with a corresponding 30 Hz PPG signal. Heart rates were chosen at random in bins of 10 bpm, and this is used as the target heart rate for augmentation. The data provided by each original video has effectively been multiplied by 10. All videos with augmented heart rates are trimmed to the same length, to prevent overfitting to the slower heart rates.

### 2.2.4 Face Detection, Cropping, and Resizing

One further augmentation of the data is to remove some unnecessary data in the background. In general, the subjects in MR-NIRP (Indoor) were closer to the camera than those in MR-NIRP (Driving). This is rectified by cropping with 25% padding to the face, detected by a non-public industry face detector. This should allow for more detail from the face to transfer into the resized images.

## 2.3 Training & Evaluation

The model is trained on an 19/7 subject train/test split, training on the augmented data but testing on the original data. For continuity, only 940 nm video is used, and the videos with motion are excluded from this experiment.

MSE is used as the loss function, however, potential improvements could be seen from frequency-aware loss functions, as MSE can severely punish phase-shifted signals. For validation, the MAE of the HR (calculated from R-R Intervals) across a whole video is used, to provide a fair comparison between this and different solutions.

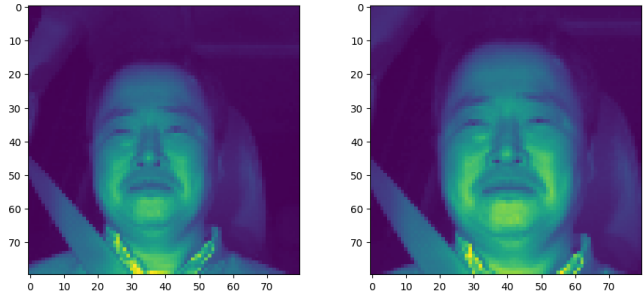

Figure 2: Uncropped vs cropped comparison

|  | Train | Test |
|---|---|---|
| **No. Subjects** | 19 | 7 |
| **No. Subjects (Indoor)** | 5 | 3 |
| **No. Subjects (Driving)** | 14 | 4 |
| **NIR Wavelengths** | 940 nm | 940 nm |
| **Motion Levels** | Still | Still |
| **Normalised PPG** | Yes | Yes |
| **Augmented HR** | Yes | No |

Table 3: Train and test sets

|  | MAE (bpm) |
|---|---|
| **Uncropped** | 1.07 |
| **Cropped** | 0.99 |

Table 4: MAE results

## 3 Results

As expected, the model performs slightly better when trained on the cropped frames, as seen in Table 4, however, the difference is more marginal than anticipated. This architecture performs better than [Nowara et al., 2021], which was trained on RGB video and tested only on the MR-NIRP (Indoor). DeepPhys [Chen and McDuff, 2018] performed better on the NIR video, however, that video focused on the neck/underside of the head, which is subject to substantially less noise than the face, while also using a much smaller dataset.

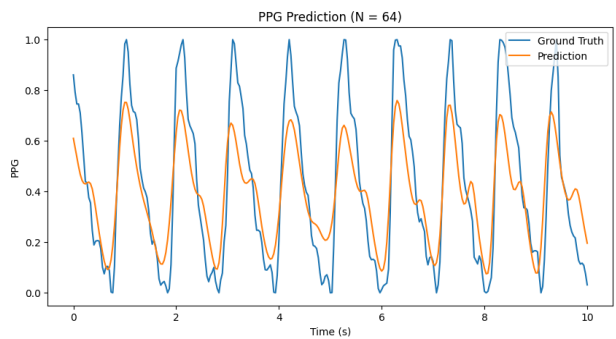

Figure 3: Example prediction

A visual inspection, as shown in Figure 3, shows that the model can consistently correctly predict peak locations, along with the respective waveforms. It also shows promise of detecting the dicrotic notch, however, it is likely that cleaner signals would be required for the model to learn these successfully.

Overall, the model shows promising signs of being able to produce accurate heart rate results while also regressing an accurate PPG signal, which may be required for further analysis.

## 4 Conclusion

Future work on this concept will include different image dimensions and different sequence lengths. The model should also be retrained on subsets of the dataset that contain the videos with motion, to improve robustness to subject movement. Further performance gains may be made by cleaning the PPG signal through band passing or other filtering methods to remove unnecessary noise. Further testing should include validation on Xperi's in-house datasets to ensure extensibility to other data.